\documentclass{article}

\usepackage[preprint]{neurips_2025}

\usepackage[utf8]{inputenc} 
\usepackage[T1]{fontenc}    
\usepackage{hyperref}       
\usepackage{url}            
\usepackage{booktabs}       
\usepackage{amsfonts}       
\usepackage{nicefrac}       
\usepackage{microtype}      
\usepackage{xcolor}         
\usepackage{graphicx}
\usepackage{subcaption}

\usepackage{cleveref}
\usepackage{enumitem}

\crefname{view}{view}{views}
\crefname{aview}{alternative view}{alternative views}
\crefname{argument}{argument}{arguments}
\crefname{cargument}{counter-argument}{counter-arguments}
\crefname{step}{step}{steps}
\crefname{barrier}{barrier}{barriers}
\crefname{wdefinition}{definition}{definitions}

\title{
    Small Language Models are the Future of Agentic AI
}

\author{%
  Peter Belcak$^{1}$~~~~
  Greg Heinrich$^{1}$~~~~
  Shizhe Diao$^{1}$~~~~
  Yonggan Fu$^{1}$~~~~
  Xin Dong$^{1}$\\
  \textbf{
  Saurav Muralidharan$^{1}$~~~~
  Yingyan Celine Lin$^{1,2}$~~~~
  Pavlo Molchanov$^{1}$
  }
\\
$^1$NVIDIA Research\quad
$^2$Georgia Institute of Technology \\
\texttt{agents-research@nvidia.com}
}

\begin{document}

\maketitle

\begin{abstract}
Large language models (LLMs) are often praised for exhibiting near-human performance on a wide range of tasks and valued for their ability to hold a general conversation.
The rise of agentic AI systems is, however, ushering in a mass of applications in which language models perform a small number of specialized tasks repetitively and with little variation.

Here we lay out the position that small language models (SLMs) are \textit{sufficiently powerful, inherently more suitable, and necessarily more economical for many invocations in agentic systems, and are therefore the future of agentic AI}.
Our argumentation is grounded in the current level of capabilities exhibited by SLMs, the common architectures of agentic systems, and the economy of LM deployment.
We further argue that in situations where general-purpose conversational abilities are essential, heterogeneous agentic systems (i.e., agents invoking multiple different models) are the natural choice.
We discuss the potential barriers for the adoption of SLMs in agentic systems and outline a general LLM-to-SLM agent conversion algorithm.

Our position\footnote{The views and positions expressed in
this paper are those of the authors and do
not necessarily reflect the views or positions of
any entities they represent.}, formulated as a value statement, highlights the significance of the operational and economic impact even a partial shift from LLMs to SLMs is to have on the AI agent industry.
We aim to stimulate the discussion on the effective use of AI resources and hope to advance the efforts to lower the costs of AI of the present day.
Calling for both contributions to and critique of our position, we commit to publishing all such correspondence at \texttt{research.nvidia.com/labs/lpr/slm-agents}.
\end{abstract}

\section{Introduction}
\label{section:introduction}
The deployment of agentic artificial intelligence is on a meteoric rise. Recent surveys show that more than a half of large IT enterprises are actively using AI agents, with 21\% having adopted just within the last year \cite{cloudera2025aiagents}. 
Aside from the users, markets also see substantial economic value in AI agents: As of late 2024, the agentic AI sector had seen more than USD 2bn in startup funding, was valued at USD 5.2bn, and was expected to grow to nearly USD 200bn by 2034 \cite{loucks2024autonomous, marketus2025agenticai}.
Put plainly, there is a growing expectation that AI agents will play a substantial role in the modern economy.


The core components powering most modern AI agents are (very) large language models \cite{masterman2024landscape,luo2025large}. It is the LLMs that provide the foundational intelligence that enables agents to make strategic decisions about when and how to use available tools, control the flow of operations needed to complete tasks, and, if necessary, to break down complex tasks into manageable subtasks and to perform reasoning for action planning and problem-solving \cite{masterman2024landscape,dairai2024llmagents}.
A typical AI agent then simply communicates with a chosen LLM API endpoint by making requests to centralized cloud infrastructure that hosts these models \cite{masterman2024landscape}.

LLM API endpoints are specifically designed to serve a large volume of diverse requests using one generalist LLM.
This operational model is deeply ingrained in the industry. In fact, it forms the foundation of substantial capital investment in the hosting cloud infrastructure -- estimated at USD 57bn \cite{colliers2025datacenter}.
It is anticipated that this operational model will remain the cornerstone of the industry and that the large initial investment will deliver returns comparable to traditional software and internet solutions within 3-4 years \cite{morganstanley2025genai}.

In this work, we recognize the dominance of the standard operational model but verbally challenge one of its aspects, namely the custom that the agents' requests to access language intelligence are -- in spite of their comparative simplicity -- handled by singleton choices of generalist LLMs. We state (\Cref{section:position}), argue (\Cref{section:position_arguments}), and defend (\Cref{section:alternative_views}) the position that the \textbf{small, rather than large, language models are the future of agentic AI}. We, however, recognize the business commitment and the now-legacy praxis that is the cause for the contrary state of the present (\Cref{section:status_quo}). In remedy, we provide an outline of a conversion algorithm for the migration of agentic applications from LLMs to SLMs (\Cref{section:conversion_algorithm}), and call for a wider discussion (\Cref{section:call_for_discussion}). If needed to concretize our stance, we attach a set of short case studies estimating the potential extent of LLM-to-SLM replacement in selected popular open-source agents (\Cref{appendix:case_studies}).

\section{Position}
\label{section:position}

\subsection{Definitions}
\label{section:definitions}

For the purpose of concretizing our position, we use the following working definitions:
\begin{enumerate}[label=\textbf{WD\arabic*},start=1]
    \item\label[wdefinition]{wdefinition:slm} A \textit{SLM} is a LM that can fit onto a common consumer electronic device and perform inference with latency sufficiently low to be practical when serving the agentic requests of one user.
    \item\label[wdefinition]{wdefinition:llm} An \textit{LLM} is a LM that is not a \textit{SLM}.
\end{enumerate}
We justify the wording of these definitions in \Cref{appendix:definitions}, but note that their choice has little bearing on the essence of our position. We note that as of 2025, we would be comfortable with considering most models below 10bn parameters in size to be SLMs.

We use the words \textit{agent} and \textit{agentic system} interchangeably, preferring the former when emphasizing the software with some agency as a whole (e.g., ``as seen in popular coding agents'') and the latter when highlighting the systems aspect of the agentic application as a sum of its components (e.g., ``not all LMs of an agentic system are replaceable by SLMs''). For brevity, we focus on LMs as the bedrock of agentic applications and do not explicitly consider vision-language models, although we note that our position and most of our arguments readily extend to vision-language models as well.

\subsection{Statement}
\textbf{We contend that SLMs are 
\begin{enumerate}[label=\textbf{V\arabic*}]
    \item\label[view]{view:power} principally \textit{sufficiently powerful} to handle language modeling errands of agentic applications;
    \item\label[view]{view:suitability} inherently \textit{more operationally suitable} for use in agentic systems than LLMs;
    \item\label[view]{view:economy} necessarily \textit{more economical} for the vast majority of LM uses in agentic systems than their general-purpose LLM counterparts by the virtue of their smaller size;
\end{enumerate}
and that on the basis of \cref{view:power,view:suitability,view:economy} SLMs are the future of agentic AI.
}

The phrasing of our position is deliberate. In its statement, we wish to convey that the described future development is ultimately a necessary consequence of the differences between SLMs and LLMs if the natural priorities are followed.
We do not make a recommendation or try to impose an obligation --- we make a statement of what we see as a faithful reflection of the community's values in this context.

\subsection{Elaboration}
We assert that the dominance of LLMs in the design of AI agents is both excessive and misaligned with the functional demands of most agentic use cases. While LLMs offer impressive generality and conversational fluency, the majority of agentic \textit{sub}tasks in deployed agentic systems are repetitive, scoped, and non-conversational—calling for models that are efficient, predictable, and inexpensive. In this context, SLMs not only suffice, but are often preferable. They offer several advantages: lower latency, reduced memory and computational requirements, and significantly lower operational costs, all while maintaining adequate task performance in constrained domains.

Our position stems from a pragmatic view of language model usage patterns within agentic architectures. These systems typically decompose complex goals into modular sub-tasks, each of which can be reliably handled by specialized or fine-tuned SLMs. We argue that insisting on LLMs for all such tasks reflects a misallocation of computational resources—one that is economically inefficient and environmentally unsustainable at scale.

Moreover, in cases where general reasoning or open-domain dialogue is essential, we advocate for heterogeneous agentic systems, where SLMs are used by default and LLMs are invoked selectively and sparingly. This modular composition — combining the precision and efficiency of SLMs with the generality of LLMs — enables the construction of agents that are both cost-effective and capable.

Ultimately, we observe that shifting the paradigm from LLM-centric to SLM-first architectures represents to many not only a technical refinement but also a Humean moral ought. As the AI community grapples with rising infrastructure costs and environmental concerns, adopting and normalizing the use of SLMs in agentic workflows can play a crucial role in promoting responsible and sustainable AI deployment.

\section{Position Arguments}
\label{section:position_arguments}
We support \cref{view:power,view:suitability,view:economy} by the following non-exclusive arguments.

\subsection{SLMs are already sufficiently powerful for use in agents}
\label{section:powerful}

\begin{enumerate}[label=\textbf{A\arabic*},start=1]
    \item\label[argument]{argument:powerful} SLMs are sufficiently powerful to take the place of LLMs in agentic systems. This argument stands in support of \cref{view:power}.
\end{enumerate}



Over the past few years, the capabilities of small language models  have advanced significantly.
Although the LM scaling laws remain observed, the scaling curve between model size and capabilities is becoming increasingly steeper, implying that the capabilities of newer small language models are much closer to those of previous large language models.
Indeed, recent advances show that well-designed small language models can meet or exceed the task performance previously attributed only to much larger models.

\begin{figure}[t]
    \centering
    \includegraphics[width=\textwidth]{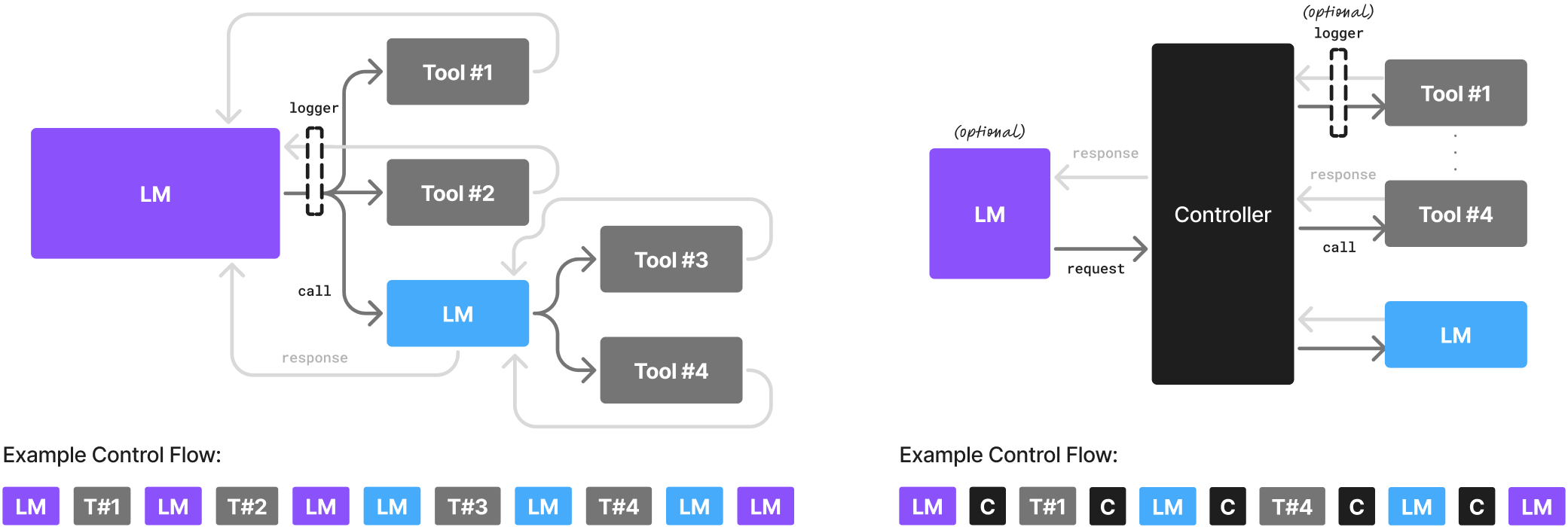}
    \caption{An illustration of agentic systems with different modes of agency. \textit{Left: Language model agency.} The language model acts both as the HCI and the orchestrator of tool calls to carry out a task. \textit{Right: Code agency.} The language model fills the role of the HCI (optionally) while a dedicated controller code orchestrates all interactions.}
    \label{figure:main}
\end{figure}

Extensive comparisons with large models have been conducted in the individual works cited below, but not all capabilities assessed by benchmarks are essential to their deployment in the agentic context. Here we highlight their aptitude for commonsense reasoning (an indicator of basic understanding), tool calling and code generation (both indicators of the ability to correctly communicate across the model→tool/code interface; see \Cref{figure:main}; \cite{berkeley-function-calling-leaderboard,yao2024tau}), and instruction following (ability to correctly respond back across the code←model interface; \cite{zhou2023instruction}). In each case, we also quote the efficiency increase if stated by the authors.

\begin{itemize}
    \item \textbf{Microsoft Phi series.} 
    Phi-2 (2.7bn) achieves commonsense reasoning scores and code generation scores on par with 30bn models while running $\sim$15$\times$ faster~\cite{phi22023}.
    Phi-3 small (7bn)~\cite{abdin2024phi} achieves language understanding and commonsense reasoning on par with and code generation scores running up to 70bn models of the same generation.
    
    \item \textbf{NVIDIA Nemotron-H family.} The 2/4.8/9bn hybrid Mamba-Transformer models achieve instruction following and code-generation accuracy comparable to dense 30bn LLMs of the same generation at an order-of-magnitude fraction of the inference FLOPs~\cite{blakeman2025nemotron}.
    
    \item \textbf{Huggingface SmolLM2 series.} SmolLM2 family of compact language models with sizes ranging from 125mn to 1.7bn parameters~\cite{allal2025smollm2smolgoesbig} each run up in their language understanding, tool calling, and instruction following performance to 14bn contemporaries while matching 70bn models of 2 years prior.
    
    \item \textbf{NVIDIA Hymba-1.5B.} This Mamba-attention hybrid-head SLM demonstrates best instruction accuracy and 3.5$\times$ greater token throughput than comparably-sized transformer models~\cite{dong2024hymba}.
    On instruction following, it outperforms larger 13bn models.
    
    \item \textbf{DeepSeek-R1-Distill series.} DeepSeek-R1-Distill is a family of reasoning models featuring 1.5-8bn sizes, trained on samples generated by DeepSeek-R1~\cite{deepseekai2025deepseekr1incentivizingreasoningcapability}. They demonstrate strong commonsense reasoning capabilities. Notably, the DeepSeek-R1-Distill-Qwen-7B model outperforms large proprietary models such as Claude-3.5-Sonnet-1022 and GPT-4o-0513.
    
    \item \textbf{DeepMind RETRO-7.5B}: Retrieval-Enhanced Transformer (RETRO) is a 7.5bn parameter model augmented with an extensive external text database, achieving performance comparable to GPT-3 (175B) on language modeling while using 25× fewer parameters~\cite{borgeaud2022retro}.
    
    \item \textbf{Salesforce xLAM-2-8B.} The 8bn model achieves state-of-the-art performance on tool calling despite is relatively modest size, surpassing frontier models like GPT-4o and Claude 3.5~\cite{zhang2024xlam}. 
\end{itemize}

Note that on top of competitive off-the-shelf performance, the reasoning capabilities of SLMs can be enhanced by light-weight selective finetuning \cite{lore2024large} or at inference time with self-consistency, verifier feedback, or tool augmentation --- e.g., Toolformer (6.7bn) outperforms GPT-3 (175bn) via API use~\cite{toolformer2023}, and appropriately prompted language models rival directly applied LLMs on math problems~\cite{zhou2022leasttomost}.

In sum, with modern training, prompting, and agentic augmentation techniques, capability --- not the parameter count --- is the binding constraint. SLMs now supply sufficient reasoning power for a substantial portion of agentic invocations, making them not just viable, but comparatively more suitable than LLMs for modular and scalable agentic systems.

\subsection{SLMs are more economical in agentic systems}
\label{section:more_economical}

\begin{enumerate}[label=\textbf{A\arabic*},start=2]
    \item\label[argument]{argument:more_economical} SLMs are more economical than LLMs in agentic systems. This argument supports \cref{view:economy}.
\end{enumerate}



Small models provide significant benefits in cost-efficiency, adaptability, and deployment flexibility. These advantages are specifically valuable in agentic workflows where specialization and iterative refinement are critical.
\Cref{section:powerful} detailed a number of efficiency comparisons of the listed SLMs to relevant LLMs. Here we draw a more encompassing picture to support \cref{argument:more_economical}.

\begin{itemize}
    \item \textbf{Inference efficiency.} Serving a 7bn SLM is 10–30$\times$ cheaper (in latency, energy consumption, and FLOPs) than a 70–175bn LLM, enabling real-time agentic responses at scale \cite{subramanian2025small,shone2024explore,invisible2025slm,mehta2024energy}. Recent advances in inference operating systems such as NVIDIA Dynamo~\cite{dynamo2025} explicitly provide support for high-throughput, low-latency SLM inference in both cloud and edge deployments. In addition, since SLMs require less or no parallelization across GPUs and nodes, the maintenance and operation of the serving infrastructure comes at a lower expense as well (see \cref{cargument:infrastructure_setup_cost} and \cref{argument:falling_setup_costs}).
    
    \item \textbf{Fine-tuning agility.} Parameter-efficient (e.g., LoRA~\cite{hu2021lora} and DoRA~\cite{liu2024dora}), low-resource \cite{belcak2025minifinetuning}, and full-parameter finetuning for SLMs require only a few GPU-hours, allowing behaviors to be added, fixed, or specialized overnight rather than over weeks \cite{subramanian2025small}.
    
    \item \textbf{Edge deployment.} Advances in on-device inference systems such as ChatRTX~\cite{chatrtx2024} demonstrate local execution of SLMs on consumer-grade GPUs, showcasing real-time, offline agentic inference with lower latency and stronger data control.
    
    \item \textbf{Parameter and embedding space utilization.} At the outset, LLMs appear to operate as monoliths involving a large amount of parameters representing swathes of compressed information in the production of their outputs. On a closer look, however, many of the embeddings passing through these systems are very sparse, engaging only a fraction of their parameters for any single input \cite{song2024powerinfer,liu2023deja} and being effectively compressible \cite{belcak2024tiny,gu2025text}. That this behavior appears to be more subdued in SLMs \cite{song2024powerinfer,xue2024powerinfer} suggests that SLMs may be fundamentally more efficient by the virtue of having a smaller proportion of their parameters contribute to the inference cost without a tangible effect on the output.
\end{itemize}

\textbf{Modular system design.} The position outlined in \cite{miehling2025agentic} presents a thorough argument in favor of composite agentic systems. Here we note that the approach of leveraging several models of varying sizes aligns well with the real-world heterogeneity of agentic tasks and is already slowly being incorporated into major software development frameworks~\cite{google-a2a}. Furthermore, this newly discovered sense for modularity in the context of agents allows for the easy addition of new skills and the ability to adapt to changing requirements, and is consistent with the push for modularity in language model design \cite{fu2024amoeballm,cai2024flextron,kudugunta2023matformer}.


The above-mentioned ``Lego-like'' composition of agentic intelligence—scaling out by adding small, specialized experts instead of scaling up monolithic models—yields systems that are cheaper, faster to debug, easier to deploy, and better aligned with the operational diversity of real-world agents. When combined with tool calling, caching, and fine-grained routing, SLM-first architectures appear to offer the best path forward for cost-effective, modular, and sustainable agentic AI.
\subsection{SLMs are more flexible}
\label{section:flexible}

\begin{enumerate}[label=\textbf{A\arabic*},start=3]
    \item\label[argument]{argument:flexible} SLMs possess greater operational flexibility in comparison to LLMs. This argument stands in support of \cref{view:suitability,view:economy}.
\end{enumerate}


Due to their small size and the associated reduction in pre-training and fine-tuning costs (\Cref{section:more_economical}), SLMs are inherently more flexible than their large counterparts when appearing in agentic systems. As such, it becomes much more affordable and practical to train, adapt, and deploy multiple specialized expert models for different agentic routines. This efficiency enables rapid iteration and adaptation, making it feasible to address evolving user needs, including supporting new behaviors, meeting new output formatting requirements, and complying with changing local regulation in selected markets \cite{wang2024comprehensive,kumar2025large,thamm2024trustworthy}.

\paragraph{Democratization.} One particularly notable and desirable consequence of SLM flexibility when put in place of LLMs is the ensuing democratization of agents. When more individuals and organizations can participate in developing language models with the aim for deployment in agentic systems, the aggregate population of agents is more likely to represent a more diverse range of perspectives and societal needs. This diversity can then help with reducing the risk of systemic biases and encourage competition and innovation. With more actors entering the field to create and refine models, the field will advance more rapidly \cite{jungherr2023artificial}.

\subsection{Agents expose only very narrow LM functionality}
\label{section:narrow_exposure}

\begin{enumerate}[label=\textbf{A\arabic*},start=4]
    \item\label[argument]{argument:narrow_exposure} Agentic applications are interfaces to a limited subset of LM capabilities. This supports \cref{view:power,view:suitability}.
\end{enumerate}

An AI agent is essentially a heavily instructed and externally choreographed gateway to a language model featuring a human-computer interface and a selection of tools that, when engaged correctly, do something of utility \cite{wang2024comprehensive}.
From this perspective, the underlying large language model that was engineered to be a powerful generalist is through a set of tediously written prompts and meticulously orchestrated context management restricted to operate within a small section of its otherwise large pallet of skills.
Thus, we argue that a SLM appropriately fine-tuned for the selected prompts would suffice while having the above-mentioned benefits of increased efficiency and greater flexibility.

It could be argued back that the careful interfacing with a generalist LLM is necessary for strong performance on the narrow task because of the LLM's better understanding of the broader language and the world (\cref{aview:llm_more_general}). This is addressed in \Cref{section:llm_more_general}.

\subsection{Agentic interactions necessitate close behavioral alignment}
\label{section:necessary_alignment}

\begin{enumerate}[label=\textbf{A\arabic*},start=5]
    \item\label[argument]{argument:necessary_alignment} Agentic interactions necessitate close behavioral alignment. This aligns with \cref{view:suitability}.
\end{enumerate}

A typical AI agent has frequent interactions with code, be it through LM tool calling or by returning output that is to be parsed by a piece of agentic code that makes the LM call \cite{masterman2024landscape}.
It is essential for the success of these interactions that the generated tool call and the generated output conform to strict formatting requirements imposed on it by the order, typing, and nature of the tool's parameters, and the expectation of the code invoking the LM, respectively.
In such cases, it becomes unnecessary for the model to handle multiple different formats (e.g. JSON/XML/Python for tool calls and XML/YAML/Markdown/Latex for output \cite{meta2025llama3_3}), as only one would be chosen for consistency across the agentic application. It is also undesirable for the model to make the occasional hallucinatory mistake and respond in a format different from that being expected by the ``code parts'' of the agentic system.
It is because of this that the SLM trained with a single formatting decision enforced during its post-training or encouraged through additional fine-tuning at a low cost is preferable over a general-purpose LLM in the context of AI agents.

\subsection{Agentic systems are naturally heterogeneous}
\label{section:naturally_heterogeneous}

\begin{enumerate}[label=\textbf{A\arabic*},start=6]
    \item\label[argument]{argument:naturally_heterogeneous} Agentic systems naturally allow for heterogeneity in the selection of models that they use. This aligns with \cref{view:suitability}.
\end{enumerate}

A language model can itself be a tool called by another language model.
Likewise, every time the agent's code invokes a language model, it can, in principle, choose any language model. This is illustrated in \Cref{figure:main}.
We argue that incorporating multiple language models of different sizes and capabilities for queries or operations of different levels of complexity offers a natural way for the introduction of SLMs.
In the context of \Cref{figure:main}-\textit{Left}, an LLM can be used for the model with the root agency, while a SLM could be used for the subordinate LM. In \Cref{figure:main}-\textit{Right}, all LMs could in principle be specialized SLMs: one for conversationality, another one for carrying out controller-defined language modeling tasks.

\subsection{Agentic interactions are natural pathways for gathering data for future improvement}
\label{section:gathering_data}

\begin{enumerate}[label=\textbf{A\arabic*},start=7]
    \item\label[argument]{argument:gathering_data} Agentic interactions are a good source for data for future model improvement. This is fundamentally supportive of \cref{view:suitability}.
\end{enumerate}

As noted in \Cref{section:narrow_exposure}, invocations of tools and language models during an agentic process are often accompanied by careful prompting that focuses the language model on delivering the narrow functionality that is required at the time. 
Each one of these invocations is itself a natural source of data for future improvement (under the necessary assumption that no non-retainable confidential data is being processed). A listener decorating the tool/model call interface can gather specialized instruction data that can later be used to produce a fine-tune an expert SLM and lower the cost of that call in the future (see \texttt{logger} in \Cref{figure:main}).
We argue that this avenue is enabled by the architecture of agents \cite{masterman2024landscape} and produces high-quality organic data (that can be further post-filtered by considering the overall success of the workflow), thus making the production of expert SLMs to stand in place of LLMs a natural step in agent deployment --- not just an auxiliary effort.

\section{Alternative Views}
\label{section:alternative_views}

The following significant alternative views have been expressed in the academic and popular literature.

\subsection{LLM generalists will always have the advantage of more general language understanding}
\label{section:llm_more_general}

\begin{enumerate}[label=\textbf{AV\arabic*},start=1]
    \item\label[aview]{aview:llm_more_general} Let $\mathcal{T}$ be a single task using general language and let $L,S$ be a large and a small language model of the same generation, respectively. The performance of $L$ on $\mathcal{T}$ will always trump that of $S$.
\end{enumerate}

This alternative view disputes \cref{view:suitability} and rests on the following counter-arguments:

\begin{enumerate}[label=\textbf{CA\arabic*},start=1]
    \item\label[cargument]{cargument:scaling_laws} There is a non-negligible body of empirical evidence of the superiority of large language models in general language understanding over small language models of the same generation. LLMs acquire their language understanding capabilities in accordance with scaling laws \cite{das2025security}. Their larger scale then enables them to demonstrate better performance across a wide array of specialized natural language tasks, including text generation, translation, and reasoning, outperforming small models trained both (a) in the same general fashion and (b) from scratch specifically for these tasks \cite{naveed2023comprehensive}.
    It can then be said that to claim otherwise is to contradict the LM scaling laws \cite{hoffmann2022training,hernandez2021scaling}.

    \item\label[cargument]{cargument:semantic_hub} Moreover, recent studies also purport that LLMs possess a ``semantic hub'' mechanism, which has been hypothesized to enable them to integrate and abstract semantic information from various modalities and languages in a generalized manner \cite{zewe2025llmsemantic}. If true, LLMs could be thought to generalize knowledge across languages and domains far more effectively than smaller models, which under the same study lack the capacity for the presence of such a hub \cite{zewe2025llmsemantic}.
    It can then argued that while small language models may be efficient for narrowly defined or highly specialized tasks, their limited scale fundamentally restricts their ability to achieve the same level of general language understanding in these specialized as LLMs because of the lack of room for the internalization of complex abstractions.
\end{enumerate}

A conclusion can be then drawn that LLM generalist models will always retain the advantage of universally better performance on language tasks, no matter how narrowly defined, over small language models of the same generation. This to be to their advantage over SLMs when deployed in agentic applications.

\paragraph{Rebuttal.}
The above alternative view is the most popularly cited belief against the use of SLMs, even when only a narrow language task needs to be performed \cite{abbyy2024smallvslarge,synergy2025smallvslarge,harrisonclarke2024llmslms,aashima2024slmvsllm}.

We believe that \cref{cargument:scaling_laws} is too limited to attack \cref{view:suitability}, namely because
\begin{enumerate}[label=\textbf{A\arabic*},start=8]
    \item\label[argument]{argument:new_architectures} Popular scaling law studies assume the model architecture to be kept constant \cite{hoffmann2022training,hernandez2021scaling} within the same generation, whereas the recent work on small language model training demonstrates that there are distinct performance benefits to considering different architectures for different model sizes \cite{dong2024hymba,blakeman2025nemotron}.

    \item\label[argument]{argument:finetuning} The flexibility of small language models (\Cref{section:flexible}) comes to the rescue. A small language model can be easily fine-tuned for the task $\mathcal{T}$ of \cref{aview:llm_more_general} to perform to the desired level of reliability. This is unaccounted for in scaling law studies.

    \item\label[argument]{argument:reasoning} Reasoning (or, more generally, test-time compute scaling; see \Cref{section:more_economical}) is significantly more affordable. A small language model, still retaining its benefits of greater cross-device agility can be reasonable expected to be scalable at inference time to the desired level of reliability.
\end{enumerate}

We also believe that \cref{cargument:semantic_hub} is too arcane to attack \cref{view:suitability} because
\begin{enumerate}[label=\textbf{A\arabic*},start=11]
    \item\label[argument]{argument:decomposition} The utility of the purported ``semantic hub'' shows itself when tasks or inputs at hand to be processed by the LM are complex. However, advanced agentic systems are either designed in their entirety or at least actively prompted to perform decompositions of complex problems and inputs \cite{masterman2024landscape,dairai2024llmagents}. Therefore, we argue to the contrary that invocations of small language models within agentic systems would be on appropriately broken-down into sub-tasks so simple that any general abstract understanding due to the hub would be of little utility.
\end{enumerate}

\subsection{LLM inference will still be cheaper because of their centralization}
\label{section:centralization}

\begin{enumerate}[label=\textbf{AV\arabic*},start=2]
    \item\label[aview]{aview:centralization} The per-token inference cost benefit of the smallness of specialized SLMs in agentic applications is dwarfed by the economy of scale.
\end{enumerate}

It could be argued that the analysis in \cref{argument:more_economical} that put forth in favor of \cref{view:economy} was ignorant of the wider business of AI model deployment:

\begin{enumerate}[label=\textbf{CA\arabic*},start=3]
    \item\label[cargument]{cargument:utilization_and_load_balancing}
    It is more difficult to fully utilize and properly balance the load for an expert SLM inference endpoint than it is for a generalist LLM endpoint \cite{subramanian2025small,evans2025llmsvsslms}.

    \item\label[cargument]{cargument:infrastructure_setup_cost}
    The costs of inference infrastructure setup combined with the costs of acquiring and retaining talent for its upkeep are often omitted in inference cost calculations but figure more prominently if the deployment of (S)LMs became the responsibility of the agent service developer.
    Early industrial reports point to considerable costs associated with these operations \cite{esg2024inferencing,chui2022stateai,seda2024cloudllm}.    
\end{enumerate}

\textbf{Acknowledgment.} We acknowledge that \cref{aview:centralization} is a valid view, with the exact economical considerations being highly case-specific. We believe that the jury is still out on \cref{aview:centralization}, but that several factors hint that \cref{view:economy} might prevail:

\begin{enumerate}[label=\textbf{A\arabic*},start=12]
    \item\label[argument]{argument:flexible_scheduling} Recent improvements in inference scheduling and large inference system modularization offer unprecedented levels of inference system flexibility in monolithic computing clusters \cite{zier2025dynamo,ai-dynamo-dynamo,mann2025dynamo}, countering the traditional stance expressed in \cref{cargument:utilization_and_load_balancing}.

    \item\label[argument]{argument:falling_setup_costs} The most recent analyses on the set-up costs of inference infrastructure indicate a consistent falling trend due to underlying technological reasons \cite{zhang2024inference,adyog2025deepseek}.
\end{enumerate}

\subsection{Equally possible worlds}
\begin{enumerate}[label=\textbf{AV\arabic*},start=3]
    \item\label[aview]{aview:equally_possible} Both the agentic world utilizing SLMs and agentic world utilizing LLMs are equally possible worlds, but the ``LLM agentic world'' has a considerable head start in terms of deployment practice and optimization, and the industry inertia already funnels efforts into innovation solely in that direction.
\end{enumerate}

\textbf{Acknowledgment.} We acknowledge \cref{aview:equally_possible} as a distinct possibility, but maintain the position that the weight of advantages described across \cref{argument:powerful,argument:more_economical,argument:flexible,argument:narrow_exposure,argument:necessary_alignment,argument:naturally_heterogeneous,argument:gathering_data} can plausibily overturn the present state of affairs.

\section{Barriers to Adoption}
\label{section:status_quo}
It would be prudent to ask oneself: If the \cref{argument:powerful,argument:more_economical,argument:flexible,argument:narrow_exposure,argument:necessary_alignment,argument:naturally_heterogeneous,argument:gathering_data} are truly compelling, why do the ever newer generations of agents seemingly just perpetuate the status quo of using generalist LLMs?

We believe the following to be among the today's main barriers to wide-spread adoption of SLMs:
\begin{enumerate}[label=\textbf{B\arabic*},start=1]
    \item\label[barrier]{barrier:investment} \textbf{Large amounts of upfront investment in centralized LLM inference infrastructure.} As detailed in \Cref{section:introduction}, large capital bets have been made on the centralized LLM inference being the leading paradigm in providing AI services in the future. As such, the industry has been much quicker at building the tools and infrastructure to that end, omitting any considerations for the possibility that more decentralized SLM or on-device inference might be equally feasible in the near future.
    \item\label[barrier]{barrier:target_benchmarks} \textbf{Use of generalist benchmarks in SLM training, design, and evaluation.} It must be pointed out that much of the work on SLM design and development follows the tracks of LLM design, focusing on the same generalist benchmarks in their development \cite{lu2024small,polo2024tinybenchmarks}. On this point, \cite{dong2024hymba} notes that if one focuses solely on benchmarks measuring the agentic utility of agents, the studied SLMs easily outperform larger models.
    \item\label[barrier]{barrier:marketing} \textbf{Lack of popular awareness.} SLMs often do not receive the level of marketing intensity and press attention LLMs do, despite their better suitability in many industrial scenarios.
\end{enumerate}

We note that \cref{barrier:investment,barrier:target_benchmarks,barrier:marketing} are practical hurdles and far from being fundamental flaws of the SLM technology in the context of agentic AI. With advanced inference scheduling systems such as Dynamo \cite{dynamo2025}, \cref{barrier:investment} is being reduced to a mere effect of inertia. \cref{barrier:target_benchmarks} is becoming increasingly recognized in the field \cite{dong2024hymba,phi22023}, and it would be natural for \cref{barrier:marketing} to fall once the economic benefits of SLM deployment in agentic applications (\cref{argument:more_economical}) are better known. With the inertia of \cref{barrier:investment} in particular, we do not endeavor to give a timeline for the retreat of these barriers or the popular adoption of SLMs.




\section{LLM-to-SLM Agent Conversion Algorithm}
\label{section:conversion_algorithm}

The very nature of agentic applications enables them to eventually switch from using LLM generalists to using SLM specialists at many of their interfaces. In the following steps, we outline an algorithm that describes one possible way to carry out the change of the underlying model painlessly.

\begin{enumerate}[label=\textbf{S\arabic*},start=1]
    \item\label[step]{step:data_collection} \textbf{Secure usage data collection.} The initial step involves deploying instrumentation to log all non-HCI agent calls, capturing input prompts, output responses, contents of individual tool calls, and optionally latency metrics for a later targeted optimization. In terms of implementation, it is the recommended practice to set up encrypted logging pipelines with role-based access controls \cite{metomic2025aiagents} and anonymize all data with respect to its origins before storage \cite{workos2025secureagents}. See \texttt{logger} in \Cref{figure:main} for an illustration.
    
    \item\label[step]{step:data_curation} \textbf{Data curation and filtering.} Begin collecting data through the pipelines of \cref{step:data_collection}. Once a satisfactory amount of data has been collected (10k-100k examples being sufficient for fine-tuning of small models as a rule of thumb \cite{agarwal2024delift,ding2023parameter}), it is necessary to remove any PII, PHI, or any other application-specific sensitive data that could cause a data leak across user accounts once used to produce a SLM specialist. Many typical varieties of sensitive data can be detected and masked or removed using popular automated tools for dataset preparation \cite{sainz2024open,radhakrishnan2023certified}. Application specific inputs (e.g. legal or internal documents) can be often be automatically paraphrased to obfuscate named entities and numerical details without compromising the general information content \cite{brennan2012adversarial,yu2024privacy,yan2024protecting}.
    
    \item\label[step]{step:task_clustering} \textbf{Task clustering.} Employ unsupervised clustering techniques on the collected prompts and agent actions to identify recurring patterns of requests or internal agent operations \cite{huang2020unsupervised,liu2005comparative,diao2025climb}. These clusters help define candidate tasks for SLM specialization. The granularity of tasks will depend on the diversity of operations; common examples include intent recognition, data extraction, summarization of specific document types, or code generation with respect to tools available to the agent.
    
    \item\label[step]{step:selection} \textbf{SLM selection.} For each identified task, select one or more candidate SLMs. Criteria for selection include the SLM's inherent capabilities (e.g., instruction following, reasoning, context window size), its performance on relevant benchmarks for the task type, its licensing, and its deployment footprint (memory, computational requirements). Models of \Cref{section:more_economical} serve as good starting candidates.

    \item\label[step]{step:slm_finetuning} \textbf{Specialized SLM fine-tuning.} For each selected task and corresponding SLM candidate, prepare a task-specific dataset from the curated data collected in \cref{step:data_curation,step:task_clustering}. Then, fine-tune the chosen SLMs on these specialized datasets. PEFT techniques such as LoRA \cite{hu2022lora} or QLoRA \cite{dettmers2023qlora} can be leveraged to reduce computational costs and memory requirements associated with fine-tuning, making the process more accessible. Full fine-tuning can also be considered if resources permit and maximal adaptation is required. In some cases, it may be beneficial to use knowledge distillation, where the specialist SLM is trained to mimic the outputs of the more powerful generalist LLM on the task-specific dataset. This can help transfer some of the more nuanced capabilities of the LLM to the SLM.

    \item\label[step]{step:iteration} \textbf{Iteration and refinement.} One may retrain the SLMs and the router model periodically with new data to maintain performance and adapt to evolving usage patterns. This forms a continuous improvement loop, returning to \cref{step:data_curation} or \cref{step:selection} as appropriate.
\end{enumerate}

\section{Call for Discussion}
\label{section:call_for_discussion}
The agentic AI industry is showing the signs of a promise to have a transformative effect on white collar work and beyond.

It is the view of the authors that any expense savings or improvements on the sustainability of AI infrastructure would act as a catalyst for this transformation, and that it is thus eminently desirable to explore all options for doing so.

We therefore call for both contributions to and critique of our position, to be directed to \texttt{agents@nvidia.com}, and commit to publishing all such correspondence at \texttt{research.nvidia.com/labs/lpr/slm-agents}.

\bibliographystyle{plain}  
\bibliography{bibliography}

\begin{thebibliography}{10}

\bibitem{aashima2024slmvsllm}
Aashima.
\newblock Small language models vs. llms: Finding the right fit for your needs, October 2024.
\newblock Accessed: 2025-05-09.

\bibitem{abbyy2024smallvslarge}
{ABBYY}.
\newblock Small language models vs. large language models, November 2024.
\newblock Accessed: 2025-05-09.

\bibitem{abdin2024phi}
Marah Abdin, Jyoti Aneja, Hany Awadalla, Ahmed Awadallah, Ammar~Ahmad Awan, Nguyen Bach, Amit Bahree, Arash Bakhtiari, Jianmin Bao, Harkirat Behl, et~al.
\newblock Phi-3 technical report: A highly capable language model locally on your phone.
\newblock {\em arXiv preprint arXiv:2404.14219}, 2024.

\bibitem{adyog2025deepseek}
{Adyog}.
\newblock The economics of ai training and inference: How deepseek broke the cost curve, February 2025.
\newblock Accessed: 2025-05-09.

\bibitem{agarwal2024delift}
Ishika Agarwal, Krishnateja Killamsetty, Lucian Popa, and Marina Danilevksy.
\newblock Delift: Data efficient language model instruction fine tuning.
\newblock {\em arXiv preprint arXiv:2411.04425}, 2024.

\bibitem{allal2025smollm2smolgoesbig}
Loubna~Ben Allal, Anton Lozhkov, Elie Bakouch, Gabriel~Martín Blázquez, Guilherme Penedo, Lewis Tunstall, Andrés Marafioti, Hynek Kydlíček, Agustín~Piqueres Lajarín, Vaibhav Srivastav, Joshua Lochner, Caleb Fahlgren, Xuan-Son Nguyen, Clémentine Fourrier, Ben Burtenshaw, Hugo Larcher, Haojun Zhao, Cyril Zakka, Mathieu Morlon, Colin Raffel, Leandro von Werra, and Thomas Wolf.
\newblock Smollm2: When smol goes big -- data-centric training of a small language model, 2025.

\bibitem{belcak2025minifinetuning}
Peter Belcak, Greg Heinrich, Jan Kautz, and Pavlo Molchanov.
\newblock Minifinetuning: Low-data generation domain adaptation through corrective self-distillation.
\newblock {\em arXiv preprint arXiv:2506.15702}, 2025.

\bibitem{belcak2024tiny}
Peter Belcak and Roger Wattenhofer.
\newblock Tiny transformers excel at sentence compression.
\newblock {\em arXiv preprint arXiv:2410.23510}, 2024.

\bibitem{blakeman2025nemotron}
Aaron Blakeman, Aarti Basant, Abhinav Khattar, Adithya Renduchintala, Akhiad Bercovich, Aleksander Ficek, Alexis Bjorlin, Ali Taghibakhshi, Amala~Sanjay Deshmukh, Ameya~Sunil Mahabaleshwarkar, et~al.
\newblock Nemotron-h: A family of accurate and efficient hybrid mamba-transformer models.
\newblock {\em arXiv preprint arXiv:2504.03624}, 2025.

\bibitem{borgeaud2022retro}
Sebastian Borgeaud, Arthur Mensch, Jordan Hoffmann, Trevor Cai, Eliza Rutherford, Katie Millican, George van~den Driessche, Bogdan Damoc, Aidan Clark, Jan Kram{\'a}r, et~al.
\newblock Improving language models by retrieving from trillions of tokens.
\newblock {\em arXiv preprint arXiv:2112.04426}, 2022.

\bibitem{brennan2012adversarial}
Michael Brennan, Sadia Afroz, and Rachel Greenstadt.
\newblock Adversarial stylometry: Circumventing authorship recognition to preserve privacy and anonymity.
\newblock {\em ACM Transactions on Information and System Security (TISSEC)}, 15(3):1--22, 2012.

\bibitem{cai2024flextron}
Ruisi Cai, Saurav Muralidharan, Greg Heinrich, Hongxu Yin, Zhangyang Wang, Jan Kautz, and Pavlo Molchanov.
\newblock Flextron: Many-in-one flexible large language model.
\newblock In {\em Proceedings of the 41st International Conference on Machine Learning (ICML 2024)}, 2024.

\bibitem{chui2022stateai}
Michael Chui, Bryce Hall, Helen Mayhew, Alex Singla, and Alexander Sukharevsky.
\newblock The state of ai in 2022—and a half decade in review, December 2022.
\newblock Accessed: 2025-05-09.

\bibitem{cloudera2025aiagents}
{Cloudera, Inc.}
\newblock 96\% of enterprises are expanding use of ai agents, according to latest data from cloudera, April 2025.
\newblock Accessed: 2025-05-08.

\bibitem{planck2018results}
Planck Collaboration et~al.
\newblock Planck 2018 results. vi. cosmological parameters.
\newblock {\em Astronomy \& Astrophysics}, 641:A6, 2020.

\bibitem{colliers2025datacenter}
Colliers.
\newblock 2025 data center marketplace: Balancing unprecedented opportunity with strategic risk.
\newblock U.s. research report, Colliers, 2025.

\bibitem{dairai2024llmagents}
{DAIR.AI}.
\newblock Llm agents, April 2024.
\newblock Accessed: 2025-05-08.

\bibitem{das2025security}
Badhan~Chandra Das, M~Hadi Amini, and Yanzhao Wu.
\newblock Security and privacy challenges of large language models: A survey.
\newblock {\em ACM Computing Surveys}, 57(6):1--39, 2025.

\bibitem{deepseekai2025deepseekr1incentivizingreasoningcapability}
DeepSeek-AI.
\newblock Deepseek-r1: Incentivizing reasoning capability in llms via reinforcement learning, 2025.

\bibitem{dettmers2023qlora}
Tim Dettmers, Artidoro Pagnoni, Ari Holtzman, and Luke Zettlemoyer.
\newblock Qlora: Efficient finetuning of quantized llms.
\newblock {\em Advances in neural information processing systems}, 36:10088--10115, 2023.

\bibitem{diao2025climb}
Shizhe Diao, Yu~Yang, Yonggan Fu, Xin Dong, Dan Su, Markus Kliegl, Zijia Chen, Peter Belcak, Yoshi Suhara, Hongxu Yin, et~al.
\newblock Climb: Clustering-based iterative data mixture bootstrapping for language model pre-training.
\newblock {\em arXiv preprint arXiv:2504.13161}, 2025.

\bibitem{ding2023parameter}
Ning Ding, Yujia Qin, Guang Yang, Fuchao Wei, Zonghan Yang, Yusheng Su, Shengding Hu, Yulin Chen, Chi-Min Chan, Weize Chen, et~al.
\newblock Parameter-efficient fine-tuning of large-scale pre-trained language models.
\newblock {\em Nature Machine Intelligence}, 5(3):220--235, 2023.

\bibitem{dong2024hymba}
Xin Dong, Yonggan Fu, Shizhe Diao, Wonmin Byeon, Zijia Chen, Ameya~Sunil Mahabaleshwarkar, Shih-Yang Liu, Matthijs Van~Keirsbilck, Min-Hung Chen, Yoshi Suhara, et~al.
\newblock Hymba: A hybrid-head architecture for small language models.
\newblock {\em arXiv preprint arXiv:2411.13676}, 2024.

\bibitem{dynamo2025}
Amr Elmeleegy et~al.
\newblock Introducing nvidia dynamo, a low-latency distributed inference framework for scaling reasoning ai models, March 2025.
\newblock NVIDIA Technical Blog.

\bibitem{evans2025llmsvsslms}
Henry Evans.
\newblock Llms vs. slms: Balancing comprehensiveness and smart resource-saving, April 2025.
\newblock Accessed: 2025-05-09.

\bibitem{ferguson2003coarse}
Barbara~A Ferguson, Timothy~A Dreisbach, Catherine~G Parks, Gregory~M Filip, and Craig~L Schmitt.
\newblock Coarse-scale population structure of pathogenic armillaria species in a mixed-conifer forest in the blue mountains of northeast oregon.
\newblock {\em Canadian Journal of Forest Research}, 33(4):612--623, 2003.

\bibitem{fu2024amoeballm}
Yonggan Fu, Zhongzhi Yu, Junwei Li, Jiayi Qian, Yongan Zhang, Xiangchi Yuan, Dachuan Shi, Roman Yakunin, and Yingyan~Celine Lin.
\newblock Amoeballm: Constructing any-shape large language models for efficient and instant deployment.
\newblock In {\em Proceedings of the 38th Annual Conference on Neural Information Processing Systems (NeurIPS 2024)}, 2024.

\bibitem{google-a2a}
google.
\newblock {GitHub - google/A2A: An open protocol enabling communication and interoperability between opaque agentic applications.}

\bibitem{gu2025text}
David Gu, Peter Belcak, and Roger Wattenhofer.
\newblock Text compression for efficient language generation.
\newblock {\em arXiv preprint arXiv:2503.11426}, 2025.

\bibitem{harrisonclarke2024llmslms}
{Harrison Clarke}.
\newblock Large language models vs. small language models, March 2024.
\newblock Accessed: 2025-05-09.

\bibitem{hernandez2021scaling}
Danny Hernandez, Jared Kaplan, Tom Henighan, and Sam McCandlish.
\newblock Scaling laws for transfer.
\newblock {\em arXiv preprint arXiv:2102.01293}, 2021.

\bibitem{hoffmann2022training}
Jordan Hoffmann, Sebastian Borgeaud, Arthur Mensch, Elena Buchatskaya, Trevor Cai, Eliza Rutherford, Diego de~Las Casas, Lisa~Anne Hendricks, Johannes Welbl, Aidan Clark, et~al.
\newblock Training compute-optimal large language models.
\newblock {\em arXiv preprint arXiv:2203.15556}, 2022.

\bibitem{hu2021lora}
Edward~J Hu, Yelong Shen, Phillip Wallis, Zeyuan Allen-Zhu, Yuanzhi Li, Shean Wang, Lu~Wang, and Weizhu Chen.
\newblock Lora: Low-rank adaptation of large language models. arxiv 2021.
\newblock {\em arXiv preprint arXiv:2106.09685}, 2021.

\bibitem{hu2022lora}
Edward~J Hu, Yelong Shen, Phillip Wallis, Zeyuan Allen-Zhu, Yuanzhi Li, Shean Wang, Lu~Wang, Weizhu Chen, et~al.
\newblock Lora: Low-rank adaptation of large language models.
\newblock {\em ICLR}, 1(2):3, 2022.

\bibitem{huang2020unsupervised}
Shaohan Huang, Furu Wei, Lei Cui, Xingxing Zhang, and Ming Zhou.
\newblock Unsupervised fine-tuning for text clustering.
\newblock In {\em Proceedings of the 28th international conference on computational linguistics}, pages 5530--5534, 2020.

\bibitem{invisible2025slm}
{Invisible Technologies}.
\newblock How small language models can outperform llms, March 2025.
\newblock Accessed: 2025-05-21.

\bibitem{phi22023}
Mojan Javaheripi and Sébastien Bubeck.
\newblock Phi-2: The surprising power of small language models, 2023.
\newblock Microsoft Research Blog.

\bibitem{jungherr2023artificial}
Andreas Jungherr.
\newblock Artificial intelligence and democracy: A conceptual framework.
\newblock {\em Social media+ society}, 9(3):20563051231186353, 2023.

\bibitem{esg2024inferencing}
Aviv Kaufmann.
\newblock Understanding the total cost of inferencing large language models.
\newblock Technical report, Enterprise Strategy Group, April 2024.
\newblock Commissioned by Dell Technologies. Accessed: 2025-05-09.

\bibitem{kudugunta2023matformer}
Sneha Kudugunta, Aditya Kusupati, Tim Dettmers, Kaifeng Chen, Inderjit Dhillon, Yulia Tsvetkov, Hannaneh Hajishirzi, Sham Kakade, Ali Farhadi, Prateek Jain, et~al.
\newblock Matformer: Nested transformer for elastic inference.
\newblock {\em arXiv preprint arXiv:2310.07707}, 2023.

\bibitem{kumar2025large}
Akshi Kumar.
\newblock From large to small: The rise of small language models (slms) in text analytics.
\newblock 2025.

\bibitem{liu2005comparative}
Luying Liu, Jianchu Kang, Jing Yu, and Zhongliang Wang.
\newblock A comparative study on unsupervised feature selection methods for text clustering.
\newblock In {\em 2005 International Conference on Natural Language Processing and Knowledge Engineering}, pages 597--601. IEEE, 2005.

\bibitem{liu2024dora}
Shih-Yang Liu, Chien-Yi Wang, Hongxu Yin, Pavlo Molchanov, Yu-Chiang~Frank Wang, Kwang-Ting Cheng, and Min-Hung Chen.
\newblock Dora: Weight-decomposed low-rank adaptation.
\newblock {\em arXiv preprint arXiv:2402.09353}, 2024.

\bibitem{liu2023deja}
Zichang Liu, Jue Wang, Tri Dao, Tianyi Zhou, Binhang Yuan, Zhao Song, Anshumali Shrivastava, Ce~Zhang, Yuandong Tian, Christopher Re, et~al.
\newblock Deja vu: Contextual sparsity for efficient llms at inference time.
\newblock In {\em International Conference on Machine Learning}, pages 22137--22176. PMLR, 2023.

\bibitem{lore2024large}
Nunzio Lore, Sepehr Ilami, and Babak Heydari.
\newblock Large model strategic thinking, small model efficiency: transferring theory of mind in large language models.
\newblock {\em arXiv preprint arXiv:2408.05241}, 2024.

\bibitem{loucks2024autonomous}
Jeff Loucks, Gillian Crossan, Baris Sarer, China Widener, and Ariane Bucaille.
\newblock Autonomous generative ai agents: Under development.
\newblock {\em Deloitte Insights}, November 2024.
\newblock Accessed: 2025-05-08.

\bibitem{lu2024small}
Zhenyan Lu, Xiang Li, Dongqi Cai, Rongjie Yi, Fangming Liu, Xiwen Zhang, Nicholas~D Lane, and Mengwei Xu.
\newblock Small language models: Survey, measurements, and insights.
\newblock {\em arXiv preprint arXiv:2409.15790}, 2024.

\bibitem{luo2025large}
Junyu Luo, Weizhi Zhang, Ye~Yuan, Yusheng Zhao, Junwei Yang, Yiyang Gu, Bohan Wu, Binqi Chen, Ziyue Qiao, Qingqing Long, et~al.
\newblock Large language model agent: A survey on methodology, applications and challenges.
\newblock {\em arXiv preprint arXiv:2503.21460}, 2025.

\bibitem{mace1981brain}
Georgina~M Mace, Paul~H Harvey, and Timothy~H Clutton-Brock.
\newblock Brain size and ecology in small mammals.
\newblock {\em Journal of Zoology}, 193(3):333--354, 1981.

\bibitem{mann2025dynamo}
Tobias Mann.
\newblock A closer look at dynamo, nvidia's 'operating system' for ai inference, March 2025.
\newblock Accessed: 2025-05-09.

\bibitem{marketus2025agenticai}
{Market.us}.
\newblock Global agentic ai market size, share analysis by product type, agent role, agent system, end user, region and companies – industry segment outlook, market assessment, competition scenario, trends and forecast 2025–2034, March 2025.
\newblock Accessed: 2025-05-08.

\bibitem{masterman2024landscape}
Tula Masterman, Sandi Besen, Mason Sawtell, and Alex Chao.
\newblock The landscape of emerging ai agent architectures for reasoning, planning, and tool calling: A survey.
\newblock {\em arXiv preprint arXiv:2404.11584}, 2024.

\bibitem{mehta2024energy}
Sourabh Mehta.
\newblock How much energy do llms consume? unveiling the power behind ai, July 2024.
\newblock Accessed: 2025-05-21.

\bibitem{meta2025llama3_3}
{Meta Platforms, Inc.}
\newblock Model cards and prompt formats: Llama 3.3, April 2025.
\newblock Accessed: 2025-05-08.

\bibitem{metomic2025aiagents}
Metomic.
\newblock Understanding ai agents \& data security, 2025.
\newblock Accessed: 2025-05-13.

\bibitem{miehling2025agentic}
Erik Miehling, Karthikeyan~Natesan Ramamurthy, Kush~R Varshney, Matthew Riemer, Djallel Bouneffouf, John~T Richards, Amit Dhurandhar, Elizabeth~M Daly, Michael Hind, Prasanna Sattigeri, et~al.
\newblock Agentic ai needs a systems theory.
\newblock {\em arXiv preprint arXiv:2503.00237}, 2025.

\bibitem{morganstanley2025genai}
{Morgan Stanley}.
\newblock Genai revenue growth and profitability, April 2025.
\newblock Accessed: 2025-05-08.

\bibitem{naveed2023comprehensive}
Humza Naveed, Asad~Ullah Khan, Shi Qiu, Muhammad Saqib, Saeed Anwar, Muhammad Usman, Naveed Akhtar, Nick Barnes, and Ajmal Mian.
\newblock A comprehensive overview of large language models.
\newblock {\em arXiv preprint arXiv:2307.06435}, 2023.

\bibitem{chatrtx2024}
NVIDIA.
\newblock Chatrtx, 2024.
\newblock NVIDIA AI Product.

\bibitem{ai-dynamo-dynamo}
{NVIDIA}.
\newblock Nvidia dynamo: A datacenter scale distributed inference serving framework.
\newblock \url{https://github.com/ai-dynamo/dynamo}, 2025.
\newblock Accessed: 2025-05-09.

\bibitem{polo2024tinybenchmarks}
Felipe~Maia Polo, Lucas Weber, Leshem Choshen, Yuekai Sun, Gongjun Xu, and Mikhail Yurochkin.
\newblock tinybenchmarks: evaluating llms with fewer examples.
\newblock {\em arXiv preprint arXiv:2402.14992}, 2024.

\bibitem{radhakrishnan2023certified}
Lakshmi Radhakrishnan, Gundolf Schenk, Kathleen Muenzen, Boris Oskotsky, Habibeh Ashouri~Choshali, Thomas Plunkett, Sharat Israni, and Atul~J Butte.
\newblock A certified de-identification system for all clinical text documents for information extraction at scale.
\newblock {\em JAMIA open}, 6(3):ooad045, 2023.

\bibitem{rees1997before}
Martin~J Rees.
\newblock {\em Before the Beginning: Our Universe and Others}.
\newblock Addison-Wesley, 1997.

\bibitem{sainz2024open}
Judith S{\'a}inz-Pardo~D{\'\i}az and {\'A}lvaro L{\'o}pez~Garc{\'\i}a.
\newblock An open source python library for anonymizing sensitive data.
\newblock {\em Scientific data}, 11(1):1289, 2024.

\bibitem{toolformer2023}
Timo Schick, Jane Dwivedi-Yu, Roberto Dessì, Roberta Raileanu, Maria Lomeli, Luke Zettlemoyer, Nicola Cancedda, and Thomas Scialom.
\newblock Toolformer: Language models can teach themselves to use tools.
\newblock In {\em Advances in Neural Information Processing Systems (NeurIPS)}, 2023.

\bibitem{schopf1993microfossils}
J~William Schopf.
\newblock Microfossils of the early archean apex chert: New evidence of the antiquity of life.
\newblock {\em Science}, 260(5108):640--646, 1993.

\bibitem{seda2024cloudllm}
Tanya Seda.
\newblock Cloud llm cost model: Breakdown for mid-market businesses, 2024.
\newblock Accessed: 2025-05-09.

\bibitem{shone2024explore}
Olivia Shone.
\newblock Explore ai models: Key differences between small language models and large language models, November 2024.
\newblock Accessed: 2025-05-21.

\bibitem{song2024powerinfer}
Yixin Song, Zeyu Mi, Haotong Xie, and Haibo Chen.
\newblock Powerinfer: Fast large language model serving with a consumer-grade gpu.
\newblock In {\em Proceedings of the ACM SIGOPS 30th Symposium on Operating Systems Principles}, pages 590--606, 2024.

\bibitem{subramanian2025small}
Shreyas Subramanian, Vikram Elango, and Mecit Gungor.
\newblock Small language models (slms) can still pack a punch: A survey.
\newblock {\em arXiv preprint arXiv:2501.05465}, 2025.

\bibitem{synergy2025smallvslarge}
{Synergy Technical}.
\newblock Small language models vs. large language models, 2025.
\newblock Accessed: 2025-05-09.

\bibitem{thamm2024trustworthy}
Brian~G. Thamm.
\newblock Trustworthy and secure ai: How small language models strengthen data security.
\newblock {\em Service Contractor Magazine}, October 2024.
\newblock Accessed: 2025-05-08.

\bibitem{wang2024comprehensive}
Fali Wang, Zhiwei Zhang, Xianren Zhang, Zongyu Wu, Tzuhao Mo, Qiuhao Lu, Wanjing Wang, Rui Li, Junjie Xu, Xianfeng Tang, et~al.
\newblock A comprehensive survey of small language models in the era of large language models: Techniques, enhancements, applications, collaboration with llms, and trustworthiness.
\newblock {\em arXiv preprint arXiv:2411.03350}, 2024.

\bibitem{workos2025secureagents}
WorkOS.
\newblock Build secure ai agents, 2025.
\newblock Accessed: 2025-05-13.

\bibitem{xue2024powerinfer}
Zhenliang Xue, Yixin Song, Zeyu Mi, Xinrui Zheng, Yubin Xia, and Haibo Chen.
\newblock Powerinfer-2: Fast large language model inference on a smartphone.
\newblock {\em arXiv preprint arXiv:2406.06282}, 2024.

\bibitem{yan2024protecting}
Biwei Yan, Kun Li, Minghui Xu, Yueyan Dong, Yue Zhang, Zhaochun Ren, and Xiuzhen Cheng.
\newblock On protecting the data privacy of large language models (llms): A survey.
\newblock {\em arXiv preprint arXiv:2403.05156}, 2024.

\bibitem{berkeley-function-calling-leaderboard}
Fanjia Yan, Huanzhi Mao, Charlie Cheng-Jie Ji, Tianjun Zhang, Shishir~G. Patil, Ion Stoica, and Joseph~E. Gonzalez.
\newblock Berkeley function calling leaderboard.
\newblock \url{https://gorilla.cs.berkeley.edu/blogs/8_berkeley_function_calling_leaderboard.html}, 2024.

\bibitem{yao2024tau}
Shunyu Yao, Noah Shinn, Pedram Razavi, and Karthik Narasimhan.
\newblock Tau-bench: A benchmark for tool-agent-user interaction in real-world domains.
\newblock {\em arXiv preprint arXiv:2406.12045}, 2024.

\bibitem{yu2024privacy}
Da~Yu, Peter Kairouz, Sewoong Oh, and Zheng Xu.
\newblock Privacy-preserving instructions for aligning large language models.
\newblock {\em arXiv preprint arXiv:2402.13659}, 2024.

\bibitem{zewe2025llmsemantic}
Adam Zewe.
\newblock Like human brains, large language models reason about diverse data in a general way.
\newblock {\em MIT News}, February 19 2025.
\newblock Accessed: 2025-05-09.

\bibitem{zhang2024xlam}
Jianguo Zhang, Tian Lan, Ming Zhu, Zuxin Liu, Thai Hoang, Shirley Kokane, Weiran Yao, Juntao Tan, Akshara Prabhakar, Haolin Chen, et~al.
\newblock xlam: A family of large action models to empower ai agent systems.
\newblock {\em arXiv preprint arXiv:2409.03215}, 2024.

\bibitem{zhang2024inference}
Kevin Zhang.
\newblock A deep dive on ai inference startups, 2024.
\newblock Accessed: 2025-05-09.

\bibitem{zhou2023instruction}
Jeffrey Zhou, Tianjian Lu, Swaroop Mishra, Siddhartha Brahma, Sujoy Basu, Yi~Luan, Denny Zhou, and Le~Hou.
\newblock Instruction-following evaluation for large language models.
\newblock {\em arXiv preprint arXiv:2311.07911}, 2023.

\bibitem{zhou2022leasttomost}
Xuezhi Zhou, Nathanael Sch{\"a}rli, Yujie Hou, Jason Wei, Denny Zhou, Quoc~V. Le, and Douwe Kiela.
\newblock Least-to-most prompting enables complex reasoning in large language models.
\newblock {\em arXiv preprint arXiv:2205.10625}, 2022.

\bibitem{zier2025dynamo}
David Zier and Harry Kim.
\newblock Introducing nvidia dynamo, a low-latency distributed inference framework for scaling reasoning ai models, March 2025.
\newblock Accessed: 2025-05-09.

\end{thebibliography}


\appendix

\clearpage
\section{Definitions}
\label{appendix:definitions}

This appendix provides two justifications for the choice of definitions in \Cref{section:definitions}.

\subsection{Pragmatic argument}
It is desirable to have a definition of SLMs that meets three key criteria:

\begin{itemize}
    \item \textbf{Timelessness.} The definition should be timeless: It should avoiding dependence on hardware-specific metrics like parameter count or FLOPs, which quickly become obsolete as technology advances—what qualifies as ``small'' today may be ``large'' tomorrow.
    \item \textbf{Practicality.} The definition is likely to have much wider generality if it is grounded in practical use, reflecting the real-world goal of deploying SLMs on widely available consumer devices, where they can serve the user in their proximity with low-latency inference.
    \item \textbf{Motivation alignment.} The definition should capture the fundamental motivation that drives the training of SLMs in the first place, which is to enable capable language models that can run on-device or within significantly constrained budgets compared to LLMs.
\end{itemize}

We find \cref{wdefinition:slm} to possess all three. \Cref{wdefinition:llm} is then phrased to complement the set of all language models.

\subsection{Limit argument}

To explore the distinction between small and large language models in the context of agentic AI, let us adopt the uncompromising lens of an extremalist, for whom intelligence must be either maximally small or maximally large.

Imagine a super-intelligent system spanning galactic scales, marshaling all available matter to optimize its computations. Such a system, while theoretically capable of addressing profound questions would face insurmountable physical constraints. The speed of light limits communication, with round-trip delays across a galaxy potentially spanning tens of thousands of years \cite{rees1997before}. This latency precludes real-time coordination, fragmenting the system into loosely coupled components rather than a unified "mind". At cosmological scales, spanning millions or billions of light-years, communication delays could approach or exceed the universe’s age of 13.8 billion years \cite{planck2018results}. Such a system, while vast, would be impractical for human-relevant applications, its computations unfolding over eons.

Conversely, consider an infinitely small intelligent system, reduced to the minimal substrate capable of computation. Such a system, akin to the simplest biological organisms, would lack the sensors, effectors, or computational capacity to meaningfully interact with its environment. Its intelligence would be constrained to rudimentary evolution, much like early life forms that emerged 3.5 billion years ago \cite{schopf1993microfossils}. Yet, even in nature, scale varies dramatically: living organisms range from bacteria (hundreds of nanometers) to blue whales (up to 30 meters), the heaviest ones being limited by heat dissipation due to their high volume-to-surface ratio \cite{ferguson2003coarse}. At the cosmic scale, all terrestrial life appears microscopic, suggesting that absolute size is less critical than functional adaptability.

Hereby: Humans, often regarded as a pinnacle of intelligence, offer a useful anchor for defining SLMs and LLMs. With a brain-to-body mass ratio surpassed only by small mammals like mice~\cite{mace1981brain}, humans balance computational efficiency with practical embodiment. SLMs, by analogy, are systems compact enough to run on personal devices, be trained with modest human interaction, or perform constrained, verifiable tasks. LLMs, in contrast, demand datacenter-scale infrastructure, organization-level training, and extensive validation, reflecting their computational load. The extremalist perspective hints at a profound truth: intelligence is not defined by size alone but by the balance of capability, efficiency, and context. For agentic workflows, SLMs may offer agility and accessibility, while LLMs provide depth at the cost of scale.

It is because of this apparent continuum that, if pressed to provide a \textit{definition} of SLMs, we choose to anchor it in characteristics of a model that can be deployed in a distributed fashion with present-day technology and be interactive enough when engaging with a human to be of utility. Proceeding in such a way, the contemporary instances of the definition will evolve as the technology underpinning these models advances, making the definition sufficiently timeless to be practical.

\section{LLM-to-SLM Replacement Case Studies}
\label{appendix:case_studies}

This appendix assesses the potential extent of replacing large language model invocations with small language models in three popular open-source agents: \textit{MetaGPT}, \textit{Open Operator}, and \textit{Cradle}. Each case study examines the use of LLMs, evaluates where SLMs may be viable replacements, and concludes with an estimated percentage of replaceable queries.

\subsection{Case study 1: MetaGPT}
\paragraph{Name.} MetaGPT

\paragraph{License.} Apache 2.0

\paragraph{Purpose.}
MetaGPT is a multi-agent framework designed to emulate a software company. It assigns roles such as Product Manager, Architect, Engineer, and QA Engineer to collaboratively handle tasks including requirement drafting, system design, implementation, and testing.

\paragraph{LLM Invocations.}
\begin{itemize}
    \item \textit{Role-Based Actions.} Each agent role invokes LLMs to fulfill its specialized responsibilities (e.g., coding, documentation).
    \item \textit{Prompt Templates.} Structured prompts used for consistent outputs.
    \item \textit{Dynamic Intelligence.} Used for planning, reasoning, and adaptation.
    \item \textit{Retrieval-Augmented Generation (RAG).} Retrieves relevant documents to enhance generation.
\end{itemize}

\paragraph{Assessment for SLM Replacement.}
SLMs would be well-suited for routine code generation and boilerplate tasks, as well as for producing structured responses based on predefined templates. However, they would require further fine-tuning data to reliably perform more complex tasks, such as architectural reasoning and adaptive planning or debugging, which would initially benefit from the broader contextual understanding and the generality of LLMs.

\paragraph{Conclusion.}
In the case of MetaGPT, we estimate that about 60\% of its LLM queries could be reliably handled by appropriately specialized SLMs.

\subsection{Case study 2: Open Operator}
\paragraph{Name.} Open Operator

\paragraph{License.} MIT License

\paragraph{Purpose.}
Open Operator is a workflow automation agent enabling users to define behaviours of agents that can perform tasks like API calls, monitoring, and orchestration using tools and services.

\paragraph{LLM Invocations}
\begin{itemize}
    \item \textit{Natural Language Processing.} Parses user intent.
    \item \textit{Decision Making.} Guides execution flow.
    \item \textit{Content Generation.} Writes summaries, reports.
\end{itemize}

\paragraph{Assessment for SLM Replacement}
SLMs would be well-suited for tasks such as simple command parsing and routing, as well as generating messages based on predefined templates. They could be meeting their limitations when dealing with more complex tasks that would require multi-step reasoning or the ability to maintain conversation flow and context over time—areas where LLMs would continue to offer significant advantages.

\paragraph{Conclusion.}
In the case of Open Operator, we estimate that about 40\% of its LLM queries could be reliably handled by appropriately specialized SLMs.

\subsection{Case study 3: Cradle}
\paragraph{Name.} Cradle

\paragraph{License.} MIT License

\paragraph{Purpose}
Cradle is designed for General Computer Control (GCC), enabling agents to operate GUI applications via screenshot input and simulated user interaction.

\paragraph{LLM Invocations.}
\begin{itemize}
    \item \textit{Interface Interpretation.} Understands visual context.
    \item \textit{Task Execution Planning.} Determines sequences of GUI actions.
    \item \textit{Error Handling.} Diagnoses and reacts to unexpected software states.
\end{itemize}

\paragraph{Assessment for SLM Replacement}
SLMs would be well-suited for handling repetitive GUI interaction workflows and the execution of pre-learned click sequences. However, they would face challenges when it comes to tasks involving dynamic GUI adaptation or unstructured error resolution, which would require a higher degree of contextual understanding typically provided by LLMs.

\paragraph{Conclusion}
In the case of Cradle, we estimate that about 70\% of its LLM queries could be reliably handled by appropriately specialized SLMs.

\end{document}